\documentclass[letterpaper]{article} 
\usepackage[submission]{aaai23}  
\usepackage{times}  
\usepackage{helvet}  
\usepackage{courier}  
\usepackage[hyphens]{url}  
\usepackage{graphicx} 
\usepackage{natbib}  
\usepackage{caption} 
\usepackage{algorithm}
\usepackage{algorithmic}
\usepackage{color}

\usepackage{newfloat}
\usepackage{listings}
\DeclareCaptionStyle{ruled}{labelfont=normalfont,labelsep=colon,strut=off} 
\floatstyle{ruled}
\newfloat{listing}{tb}{lst}{}
\floatname{listing}{Listing}

\usepackage[switch]{lineno}

\usepackage{amssymb}
\usepackage{amsmath}
\usepackage{booktabs}

\title{StyleTalk: One-shot Talking Head Generation with Controllable Speaking Styles}
\author{
    Yifeng Ma\textsuperscript{\rm 1}\thanks{Work done at Netease.},
    Suzhen Wang\textsuperscript{\rm 2},
    Zhipeng Hu\textsuperscript{\rm 2,3},
    Changjie Fan\textsuperscript{\rm 2}, \\
    Tangjie Lv\textsuperscript{\rm 2},
    Yu Ding\textsuperscript{\rm 2,3}\thanks{Corresponding authors.}, 
    Zhidong Deng\textsuperscript{\rm 1}\footnotemark[2],
    Xin Yu\textsuperscript{\rm 4} 
}
\affiliations{
    \textsuperscript{\rm 1} Department of Computer Science and Technology, 
    BNRist, THUAI, \\
State Key Laboratory of Intelligent Technology and Systems, Tsinghua University \\
    \textsuperscript{\rm 2} Virtual Human Group, Netease Fuxi AI Lab, \textsuperscript{\rm 3} Zhejiang University \\

    \textsuperscript{\rm 4} University of Technology Sydney \\

    \{wangsuzhen, zphu, fanchangjie, hzlvtangjie, dingyu01\}@corp.netease.com \\
    mayf18@mails.tsinghua.edu.cn, michael@tsinghua.edu.cn, xin.yu@uts.edu.au 
}

\usepackage{bibentry}

\begin{document}

\maketitle


\begin{abstract}
Different people speak with diverse personalized speaking styles. Although existing \emph{one-shot} talking head methods have made significant progress in lip sync, natural facial expressions, and stable head motions, they still cannot generate diverse speaking styles in the final talking head videos. To tackle this problem, we propose a one-shot style-controllable talking face generation framework. In a nutshell, we aim to attain a speaking style from an arbitrary reference speaking video and then drive the one-shot portrait to speak with the reference speaking style and another piece of audio. Specifically, we first develop a style encoder to extract dynamic facial motion patterns of a style reference video and then encode them into a style code. Afterward, we introduce a style-controllable decoder to synthesize stylized facial animations from the speech content and style code. In order to integrate the reference speaking style into generated videos, we design a style-aware adaptive transformer, which enables the encoded style code to adjust the weights of the feed-forward layers accordingly. Thanks to the style-aware adaptation mechanism, the reference speaking style can be better embedded into synthesized videos during decoding. Extensive experiments demonstrate that our method is capable of generating talking head videos with diverse speaking styles from only one portrait image and an audio clip while achieving authentic visual effects. Project Page: https://github.com/FuxiVirtualHuman/styletalk.

\end{abstract}

\begin{figure}[t]
\centering
\includegraphics[width=0.49\textwidth]{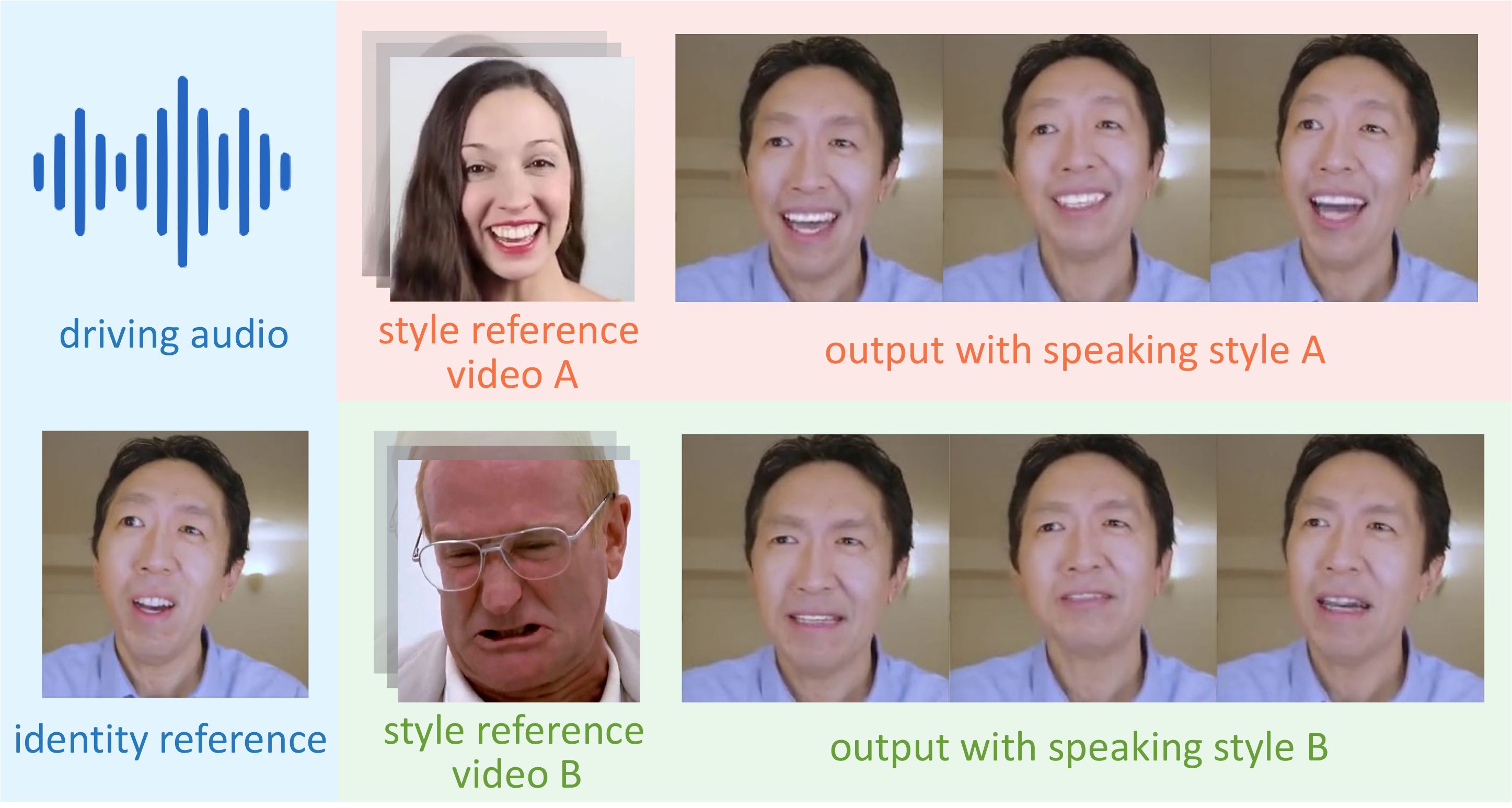}
\caption{Illustration of our proposed framework.
Given the \emph{one-shot} image of the target speaker, our approach  produces stylized photo-realistic talking faces, in which the speaker speaks the audio content with different speaking styles as shown in the additional \emph{style reference} videos. 
Note that speech content is not from the style reference video.
}
\label{fig:first_page}
\end{figure}

\begin{figure*}[ht]
\centering
\includegraphics[width=0.98\textwidth]{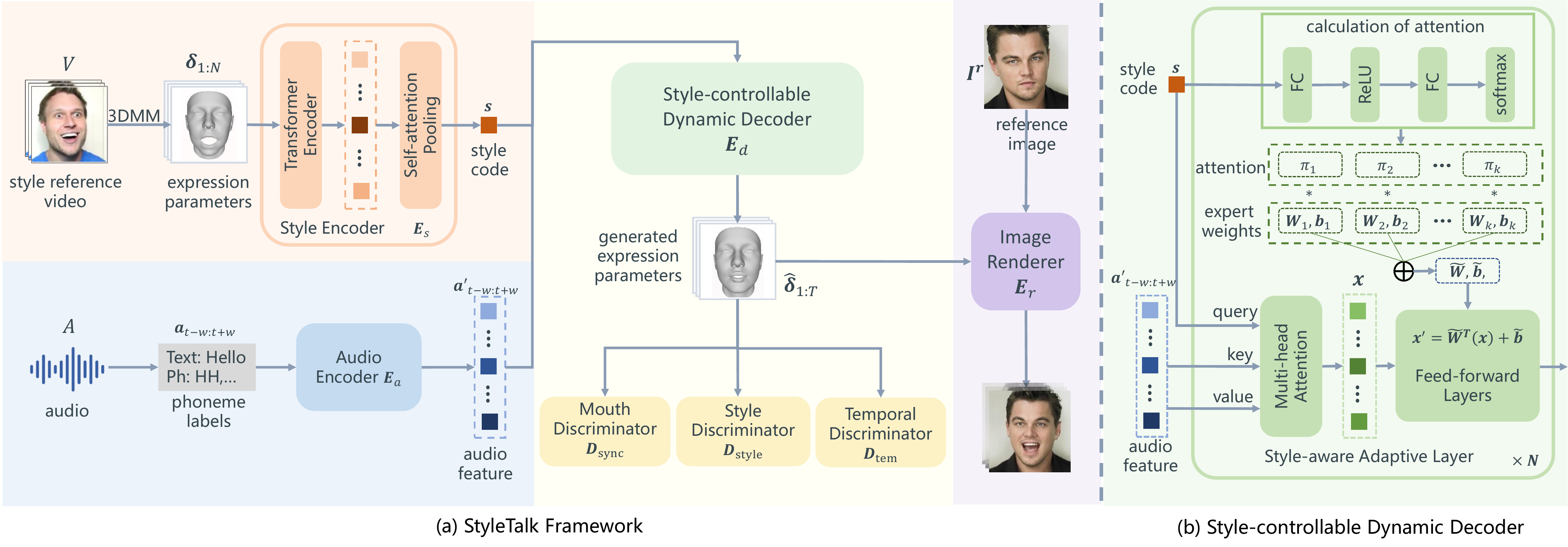}
\caption{(a) The pipeline of StyleTalk. Our framework first extracts sequential 3DMM expression parameters $\boldsymbol{\delta}_{1:N}$ from the style reference video $\boldsymbol{V}$ and then feeds them into the style encoder $\boldsymbol{E}_s$ to obtain the style code ${\boldsymbol{s}}$. An audio encoder $\boldsymbol{E}_a$ encodes phoneme labels into audio features $\boldsymbol{a}'_{t-w,t+w}$. Then the style-controllable dynamic decoder $\boldsymbol{E}_d$ generates the stylized expression parameters ${\hat{\boldsymbol{\delta}}}$ with $\boldsymbol{s}$ and $\boldsymbol{a}'$. Finally, the image renderer $\boldsymbol{E}_r$ takes the expression parameters ${\hat{\boldsymbol{\delta}}}$ and the identity reference image $\boldsymbol{I}^r$ as input and generates the output video. 
(b) The style-controllable dynamic decoder.
}
\label{fig:pipeline}
\end{figure*}

\section{Introduction}

Audio-driven photo-realistic talking head generation has drawn growing attention due to its broad applications in virtual human creation, visual dubbing, and short video creation.  The past few years have witnessed tremendous progress  in accurate lip synchronization~\cite{prajwal2020lip, wang2022one}, head pose generation~\cite{zhou2021pose, wang2021audio2head} and high-fidelity video generation~\cite{zhang2021flow, yin2022styleheat}. However, existing \emph{one-shot} based works pay less attention to modeling diverse speaking styles, thus failing to produce expressive talking head videos with various styles. 


In real-world scenarios, different people speak the same utterance with significantly diverse personalized speaking styles. Even for the same person, the speaking styles vary in different situations. Due to such significant diversities, creating style-controllable talking heads is still a great challenge, especially in the one-shot setting. In previous work~\cite{wang2020mead,sinha2022emotion}, the speaking style is merely denoted as discrete emotion classes. Such a formulation is far from representing flexible speaking styles. 
Although some recent methods \cite{ji2022eamm,liang2022expressive} can control facial expressions by involving an additional emotional source video, they mainly transfer the facial motion characteristics \emph{in a frame-by-frame fashion} without modeling the temporal dynamics of facial expressions. Therefore, a universal spatio-temporal representation of speaking styles is highly desirable.


Here, we denote speaking styles as personalized dynamic facial motion patterns. We aim to generate stylized photo-realistic talking videos for a one-shot speaker image, in which the speaker speaks the given audio content with the speaking style extracted from a style reference video (style clip). Specifically, we design a novel style-controllable talking head generation framework, called \textbf{StyleTalk}. Our framework first encodes the style clip and the input audio into the corresponding latent features, and then uses them as the input of a style-controllable dynamic decoder to obtain the stylized 3D facial animations. Finally, an image renderer \cite{ren2021pirenderer} takes the 3D facial animations and the reference image as input to generate talking faces.


To be specific, our primary goal is to obtain a universal style encoder that is able to model the facial motion patterns from arbitrary style clips. Here, we employ a transformer-based \cite{vaswani2017attention} style encoder with self-attention pooling layers \cite{safari2020self} to extract the latent style code from the sequential 3D Morphable Model (3DMM) \cite{blanz1999morphable} expression parameters of one style clip. In particular, we introduce a triplet constraint on the style code space, enabling the universal style extractor to be applicable to unseen style clips. Furthermore, we observe that the learned style codes would lie in a semantically meaningful space.

Driving a one-shot talking head with different speaking styles is also a challenging one-to-many mapping problem. Although style codes can serve as the condition and transform an ambiguous one-to-many mapping into a conditional one-to-one mapping \cite{qian2021speech}, we still observe unsatisfactory lip-sync and visual artifacts when talking faces exhibit large facial motions. To solve this issue, we propose a style-controllable dynamic transformer as our decoder. Inspired by \citet{wang2020rethinking}, we found that the feed-forward layers following the multi-head attention are of great importance to style manipulation. Hence, we propose to adaptively generate the kernel weights of the feed-forward layers based on the style code. 
Specifically, we apply an attention mechanism over $K$ kernels conditioning on style codes to modulate stylized facial expressions  of the target face adaptively. Thanks to this adaptive mechanism, our method turns the one-to-many mapping problem \cite{wang2022one} to the style-controllable one-to-one mapping in the one-shot setting, thus effectively improving the lip-sync in different styles and producing more convincing facial expressions.



Extensive experiments demonstrate that our method can generate photo-realistic talking faces with diverse speaking styles while achieving accurate lip synchronization. Our contributions are summarized as:

\begin{itemize}
    \item We propose a novel \emph{one-shot} style-controllable audio-driven talking face generation framework, which creates authentic talking videos with diverse styles from one target speaker image.
    \item We propose a universal style extractor that can effectively learn talking styles from unseen speaking style clips, thus facilitating the generation of diverse stylized talking head videos.
    \item  Benefiting from our proposed style-controllable dynamic transformer decoder, we successfully produce accurate stylized lip-sync and natural stylized facial expressions.
\end{itemize}

\section{Related Work}

\subsubsection{Audio-Driven Talking Head Generation}
With the increasing demand for virtual human creation, driving talking head with audio~\cite{zhu2021deep,chen2020comprises} has attracted considerable attention. Audio-driven methods can be classified into two categories: person-specific and person-agnostic methods. 

Person-specific methods \cite{suwajanakorn2017synthesizing,fried2019text} are only applied for speakers seen during training. Most person-specific methods~\cite{yi2020audio, thies2020neural,  song2020everybody, li2021write,lah2021lipsync3d, ji2021audio, zhang20213d, zhang2021facial} first produce 3D facial animations and then synthesize photo-realistic talking videos. Recently,  \citet{guo2021ad} and \citet{liu2022semantic} introduce neural radiance fields for high-fidelity talking head generation. 

Person-agnostic methods aim to generate talking head videos in a one-shot setting.
The early methods \cite{chung2017you,song2018talking,chen2018lip,song2018talking,zhou2019talking,chen2019hierarchical,vougioukas2019realistic,das2020speech} focus only on creating accurate mouth movements that are synchronized with the speech content.
With the development of deep learning, a number of methods \cite{wiles2018x2face,chen2020talking,zhou2020makelttalk,prajwal2020lip,zhang2021flow,wang2021audio2head,zhou2021pose,wang2022one} start to produce more natural talking faces by taking the facial expressions and head poses into consideration. 
Although the aforementioned methods can generate videos for arbitrary speakers, none of these methods is able to create expressive stylized talking head videos.


\subsubsection{Stylized Talking Head Generation}
Although the expressive facial expressions is crucial in vivid talking head generation, only a few methods~\cite{sadoughi2019speech,vougioukas2019realistic,wang2020mead,wu2021imitating,ji2021audio,sinha2022emotion,ji2022eamm,liang2022expressive} take it into consideration.
\citet{ji2021audio} extract disentangled content and emotion information from audio, and then produce videos guided by the predicted landmarks. However, determining emotions only from audio  may lead to ambiguities~\cite{ji2022eamm}, limiting the applicability of an emotional talking face model.
\citet{wang2020mead} and \citet{sinha2022emotion} create emotion-controllable talking faces by employing the explicit emotion labels as input, which drop the formulation of personalized differences in speaking styles.
\citet{ji2022eamm} and \citet{liang2022expressive} generate expressive talking head by transferring the expressions in an additional emotional source video to the target speaker frame-by-frame. 
To sum up, none of the previous works captures the spatial and temporal co-activations of facial expressions.

\begin{table*}[!htbp] 
\centering
\setlength{\tabcolsep}{1.3mm}{
\begin{tabular}{ccccccccccc}
\toprule  
&\multicolumn{5}{c}{MEAD}&\multicolumn{5}{c}{HDTF} \\
\cmidrule(r){2-6}  \cmidrule(r){7-11}
Method & SSIM$\uparrow$ & CPBD$\uparrow$ & F-LMD $\downarrow$ & M-LMD $\downarrow$ &$\text{Sync}_{conf}$$\uparrow$ & SSIM$\uparrow$  & CPBD$\uparrow$ & F-LMD $\downarrow$  &   M-LMD $\downarrow$  & $\text{Sync}_{conf}$$\uparrow$ \\
\midrule

MakeitTalk       & 0.725 & 0.106 & 3.969 & 5.324 & 2.104 & 0.593 &  0.248 & 5.084 &  4.447 & 2.563  \\
Wav2Lip   & 0.795 & \textbf{0.178} & 2.718 & 4.052 & \textbf{5.257} & 0.618  & 0.299 & 4.544 & 3.630 & 3.072  \\
PC-AVS    & 0.504 & 0.071 & 5.828 & 4.970  & 2.183 & 0.422 & 0.132 & 10.506 & 3.931 & 2.701  \\
AVCT  & 0.832 &  0.139 & 2.923 & 5.520  & 2.525 & 0.755  &  0.233 & 2.733 & 3.610 & 3.147 \\
GC-AVT & 0.340  & 0.142 & 8.039 & 7.103 & 2.417 & 0.337 &  0.296 & 10.537 & 6.206  & 2.772  \\
EAMM & 0.397 & 0.084 & 6.698 & 6.478 & 1.405 & 0.387 &  0.144 & 7.031 & 6.857  & 1.799  \\
Ground Truth  & 1   &     0.222 & 0 & 0  & 4.131 & 1     &  0.307 & 0    & 0  & 3.961  \\

\cmidrule(r){1-11}
\textbf{Ours}  & \textbf{0.837} & 0.164 & \textbf{2.122}  & \textbf{3.249} & 3.474     & \textbf{0.812}   & \textbf{0.302} & \textbf{1.941}   & \textbf{2.412} & \textbf{3.165}  \\
\bottomrule 
\end{tabular}
}
\caption{The quantitative results on MEAD and HDTF.}
\label{table:quantitive_evaluation}
\end{table*}

\begin{figure*}[ht]
\centering
\includegraphics[width=0.98\textwidth]{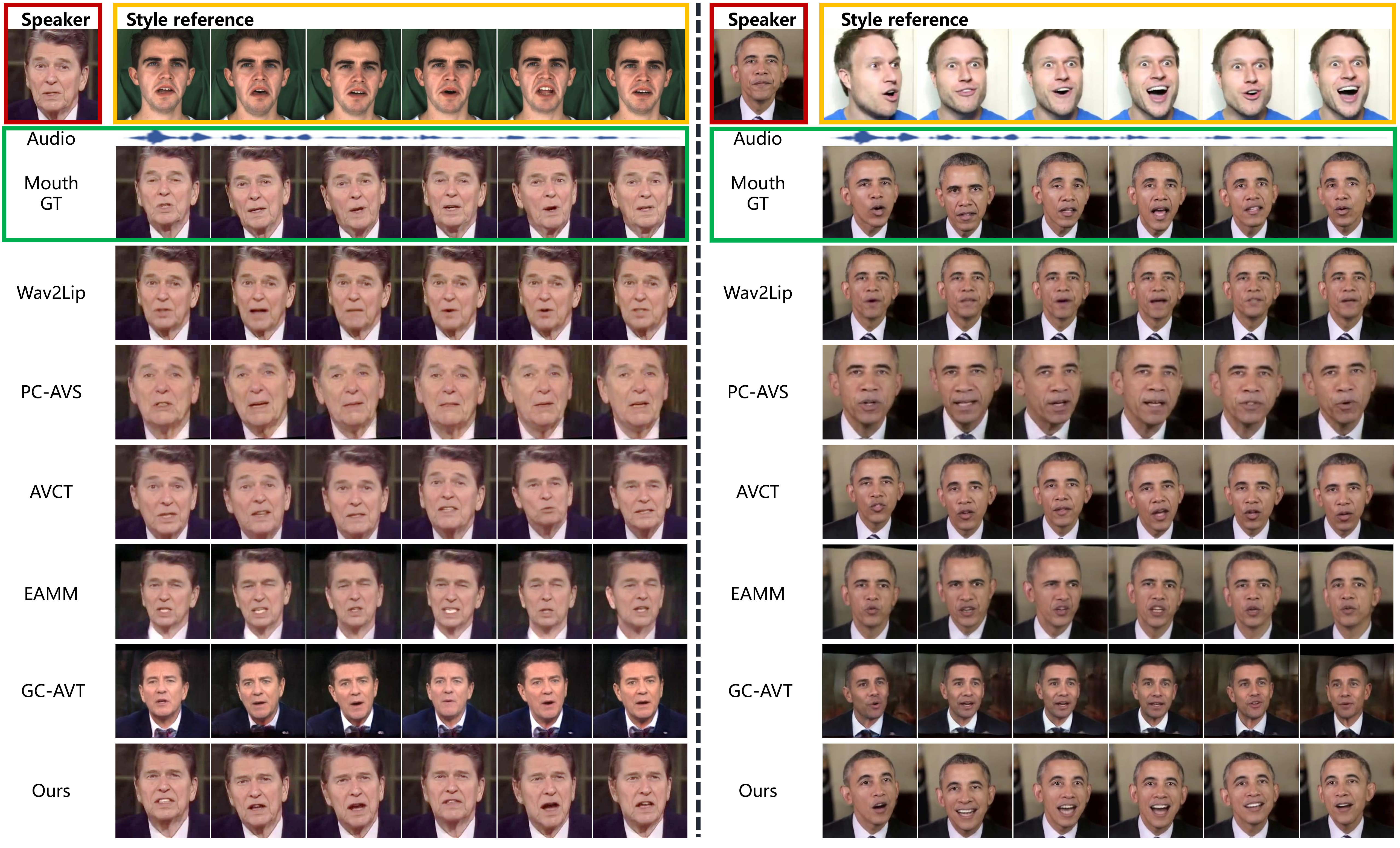}
\caption{Qualitative comparisons with the person agnostic methods. The identity reference, style reference videos and audio-synced videos are shown in the first two rows. Please zoom in or see our demo video for more details.
}
\label{fig:qualitative}
\end{figure*}

\section{Proposed Method}

In this paper, we propose a novel framework for generating the style-controllable talking faces with three inputs: (1) the reference image  $\boldsymbol{I}^r$ of the target speaker; (2)  the audio clip $\boldsymbol{A}$ of length $T$ that provides the speech content; (3) the style reference talking video $\boldsymbol{V} = \boldsymbol{I}^s_{1:N}$ of length $N$, called style clip. 
Our framework can create photo-realistic taking videos $\boldsymbol{Y} = \hat{\boldsymbol{I}}_{1:T}$ in which the target speaker speaks the speech content with the speaking style reflected in the style clip. 


As shown in Figure \ref{fig:pipeline}, 
the proposed framework consists of four components: (1) an audio encoder $\boldsymbol{E}_{a}$ which extracts the sequential pure articulation-related features $\boldsymbol{a}'_{1:T}$ from phoneme labels $\boldsymbol{a}_{1:T}$; (2) a style encoder $\boldsymbol{E}_{s}$ that encodes the facial motion patterns in the style clip into the compact style code $\boldsymbol{s}$; (3) a style-controllable dynamic decoder $\boldsymbol{E}_{d}$  which produces the stylized 3DMM expression parameters $\hat{\boldsymbol{\delta}}_{1:T}$ from the audio features and the style code; (4) an image renderer $\boldsymbol{E}_{r}$ which generates the photo-realistic talking faces using the reference image and the expression parameters. We employ PIRenderer \cite{ren2021pirenderer} as the renderer.
We adopt the training strategy proposed in \citet{wang2022one} by taking the assembled input $\{\boldsymbol{I^r},\boldsymbol{a}_{t-w,t+w},\boldsymbol{V}\}$ in a sliding window. $w$ is the window length and is set to 5.






\subsection{Audio Encoder}
The audio encoder $\boldsymbol{E}_{a}$  is expected to extract the articulation-related information from the audio. 
However, We observe that audio contains some articulation-irrelevant information, such as the emotion and the intensity, that affects the speaking style of the output.
To remove such information, we adopt the phoneme labels instead of acoustics features (e.g., Mel Frequency Cepstrum Coefficients  (MFCC)) to represent the audio signals. The phoneme labels $a_{t-w:t+w}$ are converted to word embeddings and then fed to a transformer encoder to obtain audio features $a'_{t-w:t+w}$, $a'_t \in \mathbb{R}^{256}$. The phoneme is extracted by a speech recognition tool. 

\begin{figure*}[ht]
\centering
\includegraphics[width=0.98 \textwidth]{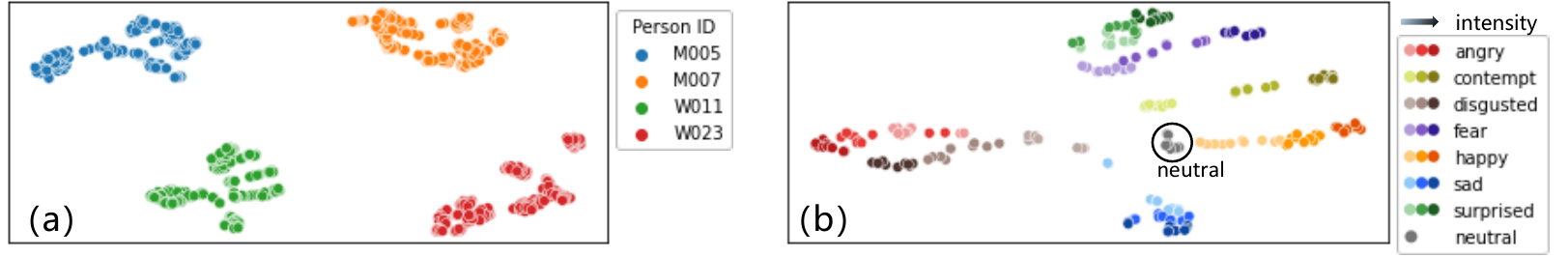}
\caption{(a) Visualization of the style codes of four speakers in MEAD. (b) Visualization of the emotional style codes of the speaker W011 in MEAD. Darker colors indicate higher emotion intensity.}
\label{fig:tsne_all}
\end{figure*}

\begin{figure}
\centering
\includegraphics[width=0.49 \textwidth]{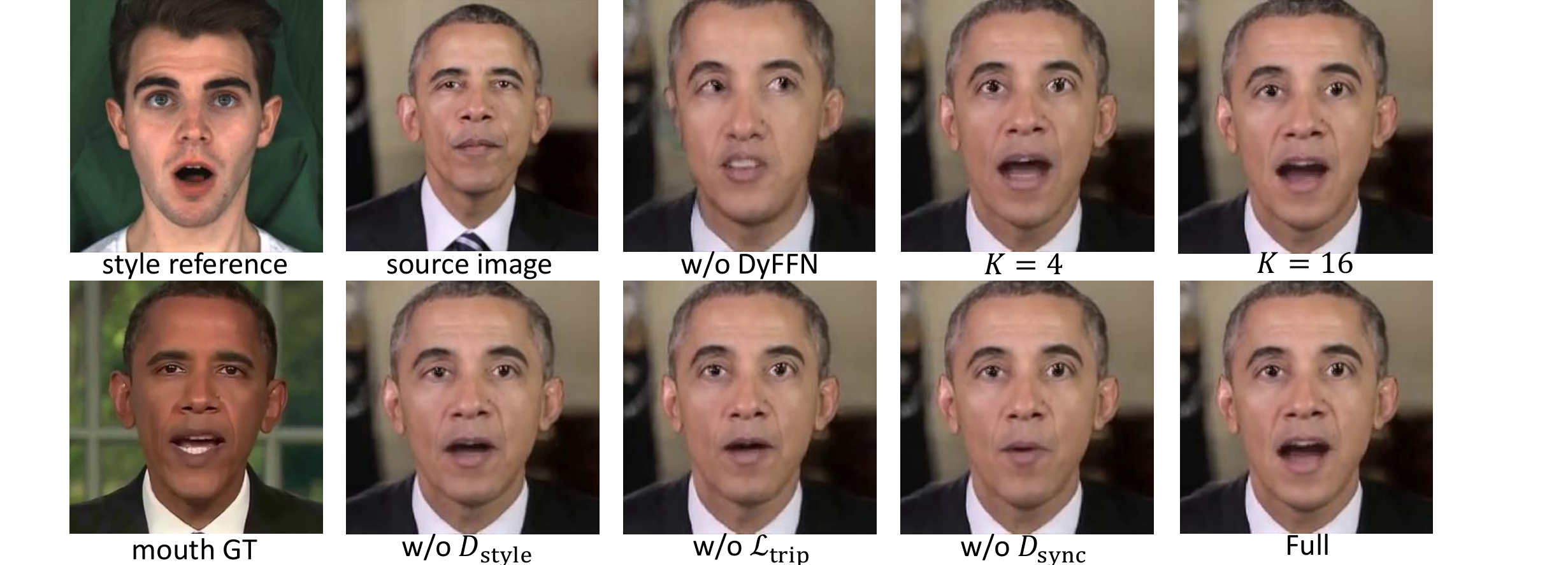}
\caption{Qualitative results of the ablation study.}
\label{fig:Qualitative_ablation}
\end{figure}

\begin{figure}
\centering
\includegraphics[width=0.49 \textwidth]{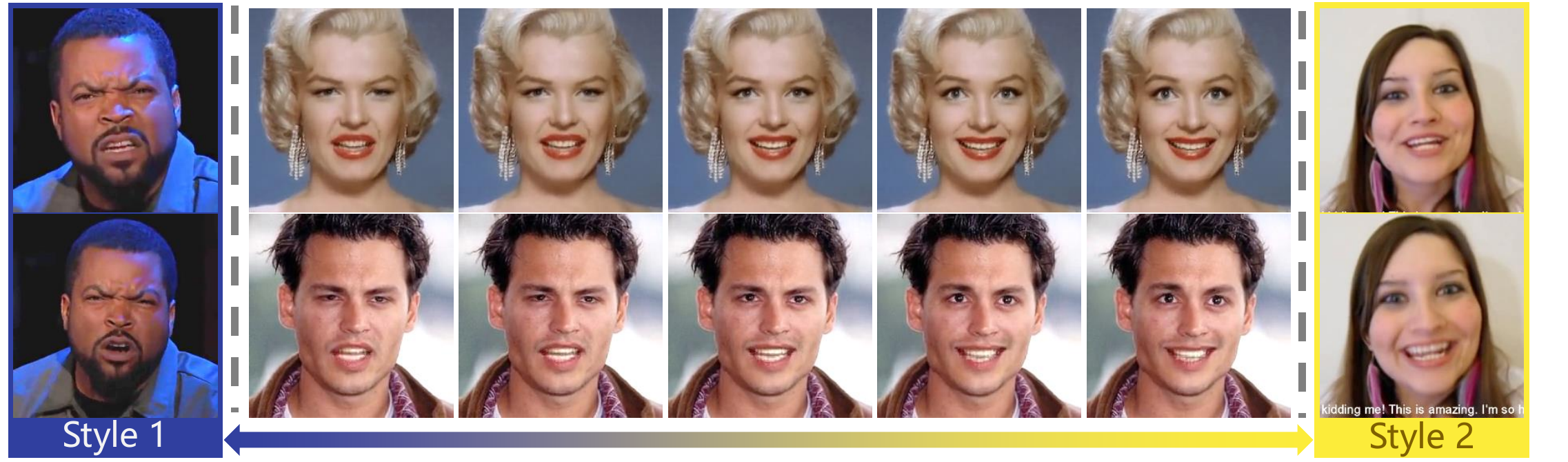}
\caption{Interpolation results between 2 speaking styles.}
\label{fig:interpolation}
\end{figure}

\subsection{Style Encoder}
The style encoder $\boldsymbol{E}_s$ extracts the speaking style reflected in the style clip. Since the speaking style is the dynamic facial motion patterns, it is irrelevant to the style clip's face shape, texture, and illumination. To remove such information, we employ the 3DMM \cite{deng2019accurate} to convert the style video clip to the the sequential expression parameters $\boldsymbol{\delta}_{1:N} \in \mathbb{R}^{N \times 64}$.

Unlike previous methods that merely transfer the static expressions of the static images \cite{ji2022eamm,liang2022expressive}, we design a style encoder to model the dynamic facial motion patterns. A transformer encoder takes the sequential 3DMM expression parameters as the input tokens. After modeling the temporal correlation between tokens, the encoder outputs the style vectors of each token, $\boldsymbol{s}'_{1:N}$. Intuitively,  the speaking style in the video clip can be identified by a few typical frames, so we employ a self-attention pooling layer \cite{safari2020self} to aggregate the style information over the style vectors. Specifically, this layer adopts an additive attention-based mechanism, which computes the token-level attention weights using a feed-forward network. The token-level attention weights represent the frame-level contributions to the video-level style code. We sum all the style vectors multiplied by the attention weights to get the final style code $\boldsymbol{s} \in \mathbb{R}^{d_s}$,
\begin{equation}
    \boldsymbol{s} = \operatorname{softmax}(W_s H)H^T,
\end{equation}
where $W_s\in \mathbb{R}^{1 \times d_s}$ is a trainable parameter, $H = [\boldsymbol{s}_1,...\boldsymbol{s}_N] \in \mathbb{R}^{d_s \times N}$ is the sequence of the encoded features, $d_s$ is the dimension of each style vector.




\subsection{Style-Controllable Dynamic Decoder}
At the early stage, we employ the vanilla transformer decoder, which takes the articulation representations $\boldsymbol{a}'_{t-w:t+w}$ and the style code $\boldsymbol{s}$ as input. Specifically, we repeat the style code $2w+1$ times and then add them with positional encodings to obtain the style tokens. The style tokens serve as the query of the transformer decoder, and the latent articulation representations serve as the key and value. The middle output token is fed into a feed-forward network to generate the output expression parameters.

When utilizing the aforementioned decoder, we observe the defective lip movements and facial expressions when generating stylized talking faces with large facial movements. Inspired by \citet{yang2019condconv} and \citet{karras2020analyzing}, we assume that the static kernel weights cannot model the  diverse speaking styles. With this assumption, we design a style-aware adaptive transformer, which dynamically adjusts the network weights according to the style code. Specifically, since \citet{wang2020rethinking} reveals that the feed-forward layers play the most important role in transformer decoder, we replace the feed-forward layers with novel style-aware adaptive feed-forward layers. 
The style-aware adaptive layer utilizes $K=8$ parallel sets of weights ${\tilde{\boldsymbol{W}}_k,\tilde{\boldsymbol{b}}_k}$.
Such parallel weights are expected to be the experts for modeling the distinct facial motion patterns of the different speaking styles.
Then we introduce the additional layers followed by Softmax to adaptively compute the attention weights over each set of weights depending on the style code. 
Then the feed-forward layer weights are aggregated dynamically via the attention weights:
\begin{equation}
    \begin{aligned}
    \tilde{\boldsymbol{W}}(\boldsymbol{s}) &=\sum_{k=1}^{K} \pi_{k}(\boldsymbol{s}) \tilde{\boldsymbol{W}}_{k}, \tilde{\boldsymbol{b}}(\boldsymbol{s})=\sum_{k=1}^{K} \pi_{k}(\boldsymbol{s}) \tilde{\boldsymbol{b}}_{k}, \\\text { s.t. } & 0 \leq \pi_{k}(\boldsymbol{s}) \leq 1, \sum_{k=1}^{K} \pi_{k}(\boldsymbol{s})=1,
    \end{aligned}
\end{equation}
where $\pi_{k}$ is the attention weight for $k^{th}$ feed-forward layer weights ${\tilde{\boldsymbol{W}}_k,\tilde{\boldsymbol{b}}_k}$. The output of style-controllable dynamic feed-forward layers is then obtained by:
\begin{equation}
    \boldsymbol{y}=g\left(\tilde{\boldsymbol{W}}^{T}(\boldsymbol{s}) \boldsymbol{x}+\tilde{\boldsymbol{b}}(\boldsymbol{s})\right),
\end{equation}
where $g$ is an activation function. Our experiments show that the style-controllable dynamic decoder helps to create accurate stylized lip movements and natural stylized facial expressions in diverse speaking styles.

\subsection{Disentanglement of Upper and Lower faces}
In experiments, we observe that the upper face and the lower face have different motion patterns. The upper face (eye, eyebrow) moves in low frequency while the lower face (mouth) moves in high frequency. Therefore, it is reasonable to model the motion patterns of the two parts with separate networks. 

We first divide expression parameters into the lower face group and the upper face group and then utilize two parallel style-controllable dynamic decoders, called the upper face decoder and the lower face decoder, to generate the corresponding group. We select 13 out of 64 expression parameters that are highly related to mouth movements as the lower face group, and the other parameters as the upper face group. The selected mouth-related PCA expression bases are reported in the supplementary materials.  The two groups of generated expression parameters are concatenated to obtain the final generated expression parameters.

\subsection{Objective Function Design}
Since our framework generates each frame individually, we adopt a batched sequential training strategy to improve the temporal consistency. Specifically, we generate successive $L=64$ frames $\boldsymbol{\delta}_{1:L}$ at one time as a clip. We then feed these frames into three discriminators: a temporal discriminator $\boldsymbol{D}_\text{tem}$,
a vertex-based lip-sync discriminator $\boldsymbol{D}_\text{sync}$,
and a style discriminator $\boldsymbol{D}_\text{style}$.
In addition, we employ the triplet constraint to obtain a semantically meaningful style space.


\subsubsection{Lip-sync Discriminator}
Because the mouth shape varies in different speaking styles, it is extremely challenging to achieve accurate lip synchronization. Inspired by \citet{prajwal2020lip}, we designed a lip-sync discriminator $D_\text{sync}$, which is trained to discriminate the synchronization between audio and mouth by randomly sampling an audio window that is either synchronous or asynchronous with a video window.
In the 3DMM, the mouth-related PCA bases also control other facial movements.
To extract pure mouth shape representation, we first convert expression parameters into face mesh using PCA bases and then pick out the mouth vertices. Instead of feeding images and acoustic features in the original SyncNet \cite{chung2016out}, we feed the mesh vertex coordinates and phonemes respectively.  We use the PointNet\cite{qi2017pointnet} as the mouth encoder to extract the mouth embedding $\mathbf{e}_m$, and a phoneme encoder to compute the audio embedding $\mathbf{e}_a$ from the phoneme window. 
We adopt cosine similarity to indicate the probability that $e_m$ and $e_a$ are synchronous:
\begin{equation}
    P_{sync} =  \frac{\mathbf{e}_m \cdot \mathbf{e}_a}{\max({\| \mathbf{e}_m \|}_2 \cdot {\| \mathbf{e}_a \|}_2, \epsilon)},
\end{equation}
where $\epsilon$ is a small constant. $D_m$ is pretrained and frozen. Our proposed framework maximize synchronous probability via a sync loss $\mathcal{L}_{\text{sync }}$on each frame of the generated clip:
\begin{equation}
    \mathcal{L}_{\text{sync }} = \frac{1}{L} \sum_{i=1}^{L}-\text{log}(P_{sync}^i).
\end{equation}

\subsubsection{Style Discriminator}
The style discriminator $\boldsymbol{D}_{\text{style}}$ is trained to determine the speaking style of the input sequential 3DMM expression parameters $\boldsymbol{\delta}_{1:L}$. 
Specifically, the style discriminator produces the probability $P^s \in \mathbb{R}^{C}$ that the sequence of parameters belongs to each speaking style. $C$ denotes the number of speaking styles. 
The style discriminator follows the structure of PatchGAN \cite{goodfellow2014generative,isola2017image,yu2017face,yu2017hallucinating,yu2018face} .The style discriminator is trained using cross-entropy loss and then frozen. The style discriminator guides the framework to generate vivid speaking styles via a style loss $\mathcal{L}_\text{style}$:
\begin{equation}
    \mathcal{L}_\text{style} = -\text{log}(P_i^s),
\end{equation}
where $i$ is the category of the ground-truth speaking style.

\subsubsection{Temporal Discriminator}
The temporal discriminator $\boldsymbol{D}_\text{tem}$ learns to distinguish the realness of the input sequences of 3DMM expression parameters $\boldsymbol{\delta}_{1:L}$. $\boldsymbol{D}_\text{tem}$ follows the structure of PatchGAN and is trained jointly with the framework by employing a GAN hinge loss $\mathcal{L}_\text{tem}$.



\subsubsection{Triplet Constraint}
Intuitively, the style codes of similar speaking styles should cluster in the style space. We employ a style triplet constraint on style codes.
Given the style clip $\boldsymbol{V}_c$ with the speaking style $c$, we randomly sample two other style clips $\boldsymbol{V}_c^p, \boldsymbol{V}_c^n$, which are and are not with the speaking style $c$, respectively. Then we obtain the corresponding style codes $\boldsymbol{s}_c$, $\boldsymbol{s}_c^p$ and $\boldsymbol{s}_c^n$. We constrain their distances in the style space with the triplet loss \cite{dong2018triplet}:
\begin{equation}
\mathcal{L}_{\text {trip }} = \max \{{\|\boldsymbol{s}_c-  \boldsymbol{s}_c^p \|}_2 - {\|\boldsymbol{s}_c-  \boldsymbol{s}_c^n \|}_2 + \gamma, 0\},
\end{equation}
where $\gamma$ is the margin parameter and is set to 5. 

\subsubsection{Total Loss}
During training, we reconstruct the facial expressions of each clip in the self-driven setting.
We adopt a combination of the L1 loss and the structural similarity (SSIM) loss \cite{wang2004image}:
\begin{equation}
    \mathcal{L}_\text{rec} = \mu \mathcal{L}_{
    \text{L1}}(\boldsymbol{\delta}_{1:L},\boldsymbol{\hat{\delta}}_{1:L}) + (1-\mu) \mathcal{L}_{\text{ssim}}(\boldsymbol{\delta}_{1:L},\boldsymbol{\hat{\delta}}_{1:L}),
\end{equation}
where $\boldsymbol{\delta}_{1:L}$ and $\boldsymbol{\hat{\delta}}_{1:L}$ are the ground truth and reconstructed facial expressions respectively. $\mu$ is a ratio coefficient and is set to 0.1. 
Our total loss is given by a combination of the aforementioned loss terms:
\begin{equation}
\begin{split}
    \mathcal{L}=\lambda_{\text {rec }} \mathcal{L}_{\text {rec }}+\lambda_{\text {trip }} \mathcal{L}_{\text {trip }}+\lambda_{\text {sync }} \mathcal{L}_{\text {sync }}+ \\
    +\lambda_{\text {tem }} \mathcal{L}_{\text {tem }}+\lambda_{\text {style }} \mathcal{L}_{\text {style }},
\end{split} 
\end{equation}
where we use $\lambda_{\text {rec }}=88$, $\lambda_{\text {trip }} = 1$, $\lambda_{\text {sync }} = 1$, $\lambda_{\text {tem }} = 1$ and $\lambda_{\text {style }} = 1$.

\begin{table}[t]
\small
\centering
\setlength{\tabcolsep}{0.5mm}{
\begin{tabular}{cccccc}
\toprule  
Method & SSIM$\uparrow$ &  CPBD$\uparrow$ & F-LMD $\downarrow$  & M-LMD $\downarrow$ & $\text{Sync}_{conf}$$\uparrow$ \\
\midrule

w/o DyFFN     & 0.830 &  \textbf{0.165} & 2.414 & 4.178 & 3.059 \\
$K=4$      & 0.831  & 0.163 & 2.327 & 3.524 & 3.331 \\
$K=16$  & 0.835  & 0.161 & 2.133 & 3.396 & 3.473 \\
w/o $D_{\text{style}}$    & 0.836  & 0.160 & 2.483   & 3.628 & 3.430  \\
w/o $\mathcal{L}_{\text{trip}}$   & \textbf{0.837}  & 0.160 & 2.401 &  3.771 & \textbf{3.532} \\
w/o $D_{\text{sync}}$  & 0.834  & 0.164 & 2.281 & 4.351 & 2.305\\
\cmidrule(r){1-6}
\textbf{Full ($K=8$)}            & \textbf{0.837}  & 0.164 & \textbf{2.122}  & \textbf{3.249} & 3.474 \\
\bottomrule 
\end{tabular}}
\caption{ Quantitive results of the ablation study on MEAD.}
\label{table:Ablation_study}
\end{table}

\section{Experiments}

\subsubsection{Datasets}
To learn a universal style extractor, we require a dataset with adequately diverse speaking styles. We construct our dataset based on the widely used datasets, MEAD \cite{wang2020mead} and HDTF \cite{zhang2021flow}.
MEAD is a in-the-lab talking-face corpus in which 60 speakers speak with three different intensity levels of eight emotions. HDTF is a high-resolution in-the-wild audio-visual dataset. 
For MEAD, we assume that the video clips, where the speaker speaks with the same emotion at the same intensity level, share the same speaking style.
For HDTF, we assume that the video clips from one speaker share the same speaking style.
Finally, we obtain 1104 speaking styles in the training set. 
The original videos are cropped and resized  to 256×256 pixels as in FOMM \cite{siarohin2019first}, and sampled at the rate of 30 FPS.

\subsubsection{Implementation Details}
Our framework is implemented by Pytorch. We employ Adam optimizer \cite{kingma2014adam} for training. $\boldsymbol{E}_r$ is trained on the combination of VoxCeleb \cite{snyder2018x}, MEAD, HDTF datasets. $\boldsymbol{D}_\text{sync}$ and $\boldsymbol{D}_\text{style}$ are trained on HDTF and MEAD for 12 hours on 4 RTX 3090 GPU with a learning rate of 0.0001. $\boldsymbol{E}_r$, $\boldsymbol{D}_\text{sync}$, $\boldsymbol{D}_\text{style}$ is then frozen. $\boldsymbol{E}_a$, $\boldsymbol{E}_s$, $\boldsymbol{E}_d$ and $\boldsymbol{D}_\text{tem}$ are trained jointly on HDTF and MEAD for 4 hours on 2 RTX 3090 GPU with a learning rate of 0.0001.

\subsection{Quantitative Evaluation}
We conduct quantitative evaluations on several widely used metrics.
To evaluate the lip synchronization, we adopt the confidence score of SyncNet~\cite{chung2016out} ($\textbf{Sync}_{\textbf{conf}}$) and the Landmark Distance around mouths (\textbf{M-LMD}) \cite{chen2019hierarchical}. To evaluate the accuracy of generated facial expressions, we adopt the Landmark Distance on the whole face (\textbf{F-LMD}). To evaluate the quality of generated talking head videos, we adopt \textbf{SSIM}, and the Cumulative Probability of Blur Detection (\textbf{CPBD}) \cite{narvekar2009no}.

We compare our method with state-of-the-art methods including: MakeitTalk \cite{zhou2020makelttalk}, Wav2Lip \cite{prajwal2020lip}, PC-AVS \cite{zhou2021pose}, AVCT \cite{wang2022one}, GC-AVT \cite{liang2022expressive}, and EAMM\cite{ji2022eamm}. We conduct the experiments in the self-driven setting on the test set, where the speaker and the speaking style are not seen during training. 
We select the first image of each video as the reference image, and the corresponding audio clip as the audio input. 
The samples of the compared methods are generated either with their released codes or with the help of their authors.
Since Wav2Lip only generates movements of the mouth area, the head poses are fixed in its samples. For other methods, poses are derived from ground truth videos.  The results of the quantitative evaluation are reported in Table \ref{table:quantitive_evaluation}.

Our method achieves the best performance among most metrics on MEAD and HDTF. Since Wav2Lip merely generates mouth movements and does not change other parts of the reference images, it obtains the highest CPBD score on MEAD. However, the mouth area generated by Wav2Lip is blurry (See Figure \ref{fig:qualitative}). Since Wav2Lip is trained using SyncNet as a discriminator, it is reasonable for Wav2Lip to obtain the highest confidence score of SyncNet on MEAD. The score is even higher than that of the ground truth. Our confidence score of SyncNet is closest to ground truth on MEAD and the highest on HDTF dataset, and our M-LMD scores are the best. This means that our method is able to achieve accurate lip-sync.
Besides, our method achieves the best performance under the F-LMD metric, which means our method is able to produce facial expressions following the reference speaking style.

\subsection{Qualitative Evaluation}
We compare our method with speaker-agnostic (one-shot) methods. The results of are displayed in Figure \ref{fig:qualitative}. The identity reference, style reference, and audio are all unseen during training. 
As can be seen, our method is able to generate talking faces with reference speaking style while achieving accurate lip-sync and preserving speaker identity 
 better(please see our demo video). 

Among all methods, only EAMM, GC-AVT, and our method achieve speaking style control. However, EAMM and GC-AVT can only control the speaking styles reflected in the upper face, i.e., eye, and eyebrow, while failing to control the stylized mouth shape. Furthermore, the speaking styles of their created videos are significantly inconsistent with those of the style reference. 
GC-AVT cannot preserve the speaking identity well. Besides, both EAMM and GC-AVT cannot produce plausible background. 

In terms of lip-sync, only Wav2Lip, AVCT, PC-AVS, and GC-AVT are competitive with our method, whereas they all can be seen as merely modeling one neutral speaking style in the mouth area, which makes achieving accurate lip-sync an easier task. EAMM cannot achieve accurate lip-sync. In contrast, our method can imitate  speaking styles in the entire face from arbitrary style clips while achieving accurate lip-sync, satisfactory identity preservation,  and producing plausible backgrounds.

We conduct a user study to further validate the effectiveness of our method and report the results in the supplementary materials.

\subsection{Ablation Study}
We conduct ablation studies on MEAD with six variants: (1) replace the adaptive feedforward layer with the vanilla feedforward layer (\textbf{w/o DyFFN}), (2)$/$(3) set  $\boldsymbol{K}=4 / 16$ in dynamic feedforward layer ($\boldsymbol{K=4}$$/$$\boldsymbol{K=16}$), 
(4) remove the style discriminator $\boldsymbol{D}_{\text{style}}$ (\textbf{w/o $\boldsymbol{D}_{\text{style}}$}),
(5) remove triplet loss (\textbf{w/o $\boldsymbol{\mathcal{L}}_{\text{trip}}$}),
(6) remove the lip-sync discriminator $\boldsymbol{D}_{\text{sync}}$ (\textbf{w/o $\boldsymbol{D}_{\text{sync}}$}), 
and (7) our full model (\textbf{Full}). The results are shown in Table \ref{table:Ablation_study} and Figure \ref{fig:Qualitative_ablation}.

Since all variants utilize the same image renderer, they obtain similar SSIM and CPBD scores. 
The variant \textbf{w/o DyFFN} obtain worse F-LMD, M-LMD and $\text{Sync}_{conf}$ scores than the \textbf{Full} model, which demonstrates the effectiveness of proposed style-aware dynamic decoder in modeling stylized facial motions. We empirically observe that \textbf{$K=8$} is the best setting for our task.
Without $\boldsymbol{D}_{\text{style}}$ and $\boldsymbol{\mathcal{L}}_{\text{trip}}$, the scores of F-LMD and M-LMD also drop dramatically. This implies that the style discriminator and the triplet constraint compel our framework to better perceive the stylized facial motion patterns.
Without the supervision of $\boldsymbol{D}_{\text{sync}}$, the results show bad lip synchronization.


\subsection{Style Space Inspection}
\subsubsection{Style Space Visualization}

We project the style codes to a 2D space using t-distributed stochastic neighbour embedding (t-SNE) \cite{van2008visualizing} . For clarity, we select the speaking styles of 4 speakers from the MEAD dataset. Each speaker has 22 speaking styles (7 emotions $\times$ 3 levels plus one neutral style). For each speaking style, we randomly select 10 video clips to extract style codes.

In figure \ref{fig:tsne_all}~(a), each speaker is marked with a distinct color. As shown, the style codes of the same speaker cluster in the style space. This implies that the speaking styles of one speaker are more similar than those with the same emotion.
Figure \ref{fig:tsne_all}~(b) shows the style codes from one speaker in MEAD dataset. Each style code is marked with a color corresponding to its emotion and intensity. Each group of style codes with the same emotion first gathers into one cluster. In each cluster, the style codes of emotions with low intensity are close to those of the neutral emotion. Notably, some emotions show similar facial motion patterns, such as anger vs disgust and surprise vs fear. Thus, their style codes are close in the style space.
The aforementioned observations prove that our model is able to learn a semantically meaningful style space.

\subsubsection{Style Manipulation}
Thanks to the meaningful style space, our method can edit the speaking styles by manipulating style codes. As shown in Figure \ref{fig:interpolation}, when linearly interpolating between two style codes extracted from unseen style clips, the speaking styles of generated videos transition smoothly.
Through interpolation, our method is able to  control the style intensity (by interpolating the style with a neutral style) and create new speaking styles. 


\section{Conclusion}

In this paper, we propose a novel talking head generation framework, \textit{StyleTalk}, which generates one-shot audio-driven talking faces with diverse speaking styles.
Our method effectively extracts the speaking style from an arbitrary style reference video and then injects the style into the facial animations of the target speaker using our proposed style-controllable modules. 
In contrast to previous works, our method captures the spatio-temporal co-activations of the facial expressions from the style reference videos, thus leading to authentic stylized talking face videos.
Extensive experiments demonstrate that our method creates photo-realistic talking head videos with a conditional speaking style while achieving more accurate lip-sync and better identity-preservation compared with the state-of-the-art.

\section{Acknowledgments}
This work is supported by the 2022 Hangzhou Key Science and Technology Innovation Program (No. 2022AIZD0054) and the Key Research and Development Program of Zhejiang Province (No. 2022C01011). This research is partially funded by the ARC-Discovery grants (DP220100800) and ARC-DECRA (DE230100477). This work was supported in part by the National Science Foundation of China (NSFC) under Grant No. 62176134, by a grant from the Institute Guo Qiang (2019GQG0002), Tsinghua University, and by research and application on AI technologies for smart mobility funded by SAIC Motor.

We would like to thank Xinya Ji, Borong Liang, Yan Pan for their generous help with the comparisons. We would also like to thank Lincheng Li and Zhimeng Zhang for helpful discussions.

\bibliography{main}

\begin{thebibliography}{59}
\providecommand{\natexlab}[1]{#1}

\bibitem[{Blanz and Vetter(1999)}]{blanz1999morphable}
Blanz, V.; and Vetter, T. 1999.
\newblock A morphable model for the synthesis of 3D faces.
\newblock In \emph{Proceedings of the 26th annual conference on Computer
  graphics and interactive techniques}, 187--194.

\bibitem[{Chen et~al.(2020{\natexlab{a}})Chen, Cui, Kou, Zheng, and
  Xu}]{chen2020comprises}
Chen, L.; Cui, G.; Kou, Z.; Zheng, H.; and Xu, C. 2020{\natexlab{a}}.
\newblock What comprises a good talking-head video generation?: A survey and
  benchmark.
\newblock \emph{arXiv preprint arXiv:2005.03201}.

\bibitem[{Chen et~al.(2020{\natexlab{b}})Chen, Cui, Liu, Li, Kou, Xu, and
  Xu}]{chen2020talking}
Chen, L.; Cui, G.; Liu, C.; Li, Z.; Kou, Z.; Xu, Y.; and Xu, C.
  2020{\natexlab{b}}.
\newblock Talking-head generation with rhythmic head motion.
\newblock In \emph{ECCV}, 35--51. Springer.

\bibitem[{Chen et~al.(2018)Chen, Li, Maddox, Duan, and Xu}]{chen2018lip}
Chen, L.; Li, Z.; Maddox, R.~K.; Duan, Z.; and Xu, C. 2018.
\newblock Lip movements generation at a glance.
\newblock In \emph{ECCV}, 520--535.

\bibitem[{Chen et~al.(2019)Chen, Maddox, Duan, and Xu}]{chen2019hierarchical}
Chen, L.; Maddox, R.~K.; Duan, Z.; and Xu, C. 2019.
\newblock Hierarchical cross-modal talking face generation with dynamic
  pixel-wise loss.
\newblock In \emph{CVPR}, 7832--7841.

\bibitem[{Chung, Jamaludin, and Zisserman(2017)}]{chung2017you}
Chung, J.~S.; Jamaludin, A.; and Zisserman, A. 2017.
\newblock You said that?
\newblock \emph{arXiv preprint arXiv:1705.02966}.

\bibitem[{Chung and Zisserman(2016)}]{chung2016out}
Chung, J.~S.; and Zisserman, A. 2016.
\newblock Out of time: automated lip sync in the wild.
\newblock In \emph{Asian conference on computer vision}, 251--263. Springer.

\bibitem[{Das et~al.(2020)Das, Biswas, Sinha, and Bhowmick}]{das2020speech}
Das, D.; Biswas, S.; Sinha, S.; and Bhowmick, B. 2020.
\newblock Speech-driven facial animation using cascaded gans for learning of
  motion and texture.
\newblock In \emph{ECCV}, 408--424. Springer.

\bibitem[{Deng et~al.(2019)Deng, Yang, Xu, Chen, Jia, and
  Tong}]{deng2019accurate}
Deng, Y.; Yang, J.; Xu, S.; Chen, D.; Jia, Y.; and Tong, X. 2019.
\newblock Accurate 3d face reconstruction with weakly-supervised learning: From
  single image to image set.
\newblock In \emph{CVPRW}, 0--0.

\bibitem[{Dong and Shen(2018)}]{dong2018triplet}
Dong, X.; and Shen, J. 2018.
\newblock Triplet loss in siamese network for object tracking.
\newblock In \emph{ECCV}, 459--474.

\bibitem[{Fried et~al.(2019)Fried, Tewari, Zollh{\"o}fer, Finkelstein,
  Shechtman, Goldman, Genova, Jin, Theobalt, and Agrawala}]{fried2019text}
Fried, O.; Tewari, A.; Zollh{\"o}fer, M.; Finkelstein, A.; Shechtman, E.;
  Goldman, D.~B.; Genova, K.; Jin, Z.; Theobalt, C.; and Agrawala, M. 2019.
\newblock Text-based editing of talking-head video.
\newblock \emph{ACM Transactions on Graphics (TOG)}, 38(4): 1--14.

\bibitem[{Goodfellow et~al.(2014)Goodfellow, Pouget-Abadie, Mirza, Xu,
  Warde-Farley, Ozair, Courville, and Bengio}]{goodfellow2014generative}
Goodfellow, I.; Pouget-Abadie, J.; Mirza, M.; Xu, B.; Warde-Farley, D.; Ozair,
  S.; Courville, A.; and Bengio, Y. 2014.
\newblock Generative adversarial nets.
\newblock \emph{Advances in neural information processing systems}, 27.

\bibitem[{Guo et~al.(2021)Guo, Chen, Liang, Liu, Bao, and Zhang}]{guo2021ad}
Guo, Y.; Chen, K.; Liang, S.; Liu, Y.; Bao, H.; and Zhang, J. 2021.
\newblock AD-NeRF: Audio Driven Neural Radiance Fields for Talking Head
  Synthesis.
\newblock \emph{arXiv preprint arXiv:2103.11078}.

\bibitem[{Isola et~al.(2017)Isola, Zhu, Zhou, and Efros}]{isola2017image}
Isola, P.; Zhu, J.-Y.; Zhou, T.; and Efros, A.~A. 2017.
\newblock Image-to-image translation with conditional adversarial networks.
\newblock In \emph{Proceedings of the IEEE conference on computer vision and
  pattern recognition}, 1125--1134.

\bibitem[{Ji et~al.(2022)Ji, Zhou, Wang, Wu, Wu, Xu, and Cao}]{ji2022eamm}
Ji, X.; Zhou, H.; Wang, K.; Wu, Q.; Wu, W.; Xu, F.; and Cao, X. 2022.
\newblock EAMM: One-Shot Emotional Talking Face via Audio-Based Emotion-Aware
  Motion Model.
\newblock \emph{arXiv preprint arXiv:2205.15278}.

\bibitem[{Ji et~al.(2021)Ji, Zhou, Wang, Wu, Loy, Cao, and Xu}]{ji2021audio}
Ji, X.; Zhou, H.; Wang, K.; Wu, W.; Loy, C.~C.; Cao, X.; and Xu, F. 2021.
\newblock Audio-driven emotional video portraits.
\newblock In \emph{CVPR}, 14080--14089.

\bibitem[{Karras et~al.(2020)Karras, Laine, Aittala, Hellsten, Lehtinen, and
  Aila}]{karras2020analyzing}
Karras, T.; Laine, S.; Aittala, M.; Hellsten, J.; Lehtinen, J.; and Aila, T.
  2020.
\newblock Analyzing and improving the image quality of stylegan.
\newblock In \emph{CVPR}, 8110--8119.

\bibitem[{Kingma and Ba(2014)}]{kingma2014adam}
Kingma, D.~P.; and Ba, J. 2014.
\newblock Adam: A method for stochastic optimization.
\newblock \emph{arXiv preprint arXiv:1412.6980}.

\bibitem[{Lahiri et~al.(2021)Lahiri, Kwatra, Frueh, Lewis, and
  Bregler}]{lah2021lipsync3d}
Lahiri, A.; Kwatra, V.; Frueh, C.; Lewis, J.; and Bregler, C. 2021.
\newblock LipSync3D: Data-Efficient Learning of Personalized 3D Talking Faces
  from Video using Pose and Lighting Normalization.
\newblock In \emph{CVPR}, 2755--2764.

\bibitem[{Li et~al.(2021)Li, Wang, Zhang, Ding, Zheng, Yu, and
  Fan}]{li2021write}
Li, L.; Wang, S.; Zhang, Z.; Ding, Y.; Zheng, Y.; Yu, X.; and Fan, C. 2021.
\newblock Write-a-speaker: Text-based Emotional and Rhythmic Talking-head
  Generation.
\newblock In \emph{AAAI}, volume~35, 1911--1920.

\bibitem[{Liang et~al.(2022)Liang, Pan, Guo, Zhou, Hong, Han, Han, Liu, Ding,
  and Wang}]{liang2022expressive}
Liang, B.; Pan, Y.; Guo, Z.; Zhou, H.; Hong, Z.; Han, X.; Han, J.; Liu, J.;
  Ding, E.; and Wang, J. 2022.
\newblock Expressive talking head generation with granular audio-visual
  control.
\newblock In \emph{CVPR}, 3387--3396.

\bibitem[{Liu et~al.(2022)Liu, Xu, Wu, Zhou, Wu, and Zhou}]{liu2022semantic}
Liu, X.; Xu, Y.; Wu, Q.; Zhou, H.; Wu, W.; and Zhou, B. 2022.
\newblock Semantic-aware implicit neural audio-driven video portrait
  generation.
\newblock \emph{arXiv preprint arXiv:2201.07786}.

\bibitem[{Narvekar and Karam(2009)}]{narvekar2009no}
Narvekar, N.~D.; and Karam, L.~J. 2009.
\newblock A no-reference perceptual image sharpness metric based on a
  cumulative probability of blur detection.
\newblock In \emph{2009 International Workshop on Quality of Multimedia
  Experience}, 87--91. IEEE.

\bibitem[{Prajwal et~al.(2020)Prajwal, Mukhopadhyay, Namboodiri, and
  Jawahar}]{prajwal2020lip}
Prajwal, K.; Mukhopadhyay, R.; Namboodiri, V.~P.; and Jawahar, C. 2020.
\newblock A lip sync expert is all you need for speech to lip generation in the
  wild.
\newblock In \emph{Proceedings of the 28th ACM International Conference on
  Multimedia}, 484--492.

\bibitem[{Qi et~al.(2017)Qi, Su, Mo, and Guibas}]{qi2017pointnet}
Qi, C.~R.; Su, H.; Mo, K.; and Guibas, L.~J. 2017.
\newblock Pointnet: Deep learning on point sets for 3d classification and
  segmentation.
\newblock In \emph{CVPR}, 652--660.

\bibitem[{Qian et~al.(2021)Qian, Tu, Zhi, Liu, and Gao}]{qian2021speech}
Qian, S.; Tu, Z.; Zhi, Y.; Liu, W.; and Gao, S. 2021.
\newblock Speech drives templates: Co-speech gesture synthesis with learned
  templates.
\newblock In \emph{ICCV}, 11077--11086.

\bibitem[{Ren et~al.(2021)Ren, Li, Chen, Li, and Liu}]{ren2021pirenderer}
Ren, Y.; Li, G.; Chen, Y.; Li, T.~H.; and Liu, S. 2021.
\newblock Pirenderer: Controllable portrait image generation via semantic
  neural rendering.
\newblock In \emph{ICCV}, 13759--13768.

\bibitem[{Sadoughi and Busso(2019)}]{sadoughi2019speech}
Sadoughi, N.; and Busso, C. 2019.
\newblock Speech-driven expressive talking lips with conditional sequential
  generative adversarial networks.
\newblock \emph{IEEE Transactions on Affective Computing}, 12(4): 1031--1044.

\bibitem[{Safari, India, and Hernando(2020)}]{safari2020self}
Safari, P.; India, M.; and Hernando, J. 2020.
\newblock Self-attention encoding and pooling for speaker recognition.
\newblock \emph{arXiv preprint arXiv:2008.01077}.

\bibitem[{Siarohin et~al.(2019)Siarohin, Lathuili{\`e}re, Tulyakov, Ricci, and
  Sebe}]{siarohin2019first}
Siarohin, A.; Lathuili{\`e}re, S.; Tulyakov, S.; Ricci, E.; and Sebe, N. 2019.
\newblock First order motion model for image animation.
\newblock \emph{Advances in Neural Information Processing Systems}, 32:
  7137--7147.

\bibitem[{Sinha et~al.(2022)Sinha, Biswas, Yadav, and
  Bhowmick}]{sinha2022emotion}
Sinha, S.; Biswas, S.; Yadav, R.; and Bhowmick, B. 2022.
\newblock Emotion-Controllable Generalized Talking Face Generation.
\newblock \emph{arXiv preprint arXiv:2205.01155}.

\bibitem[{Snyder et~al.(2018)Snyder, Garcia-Romero, Sell, Povey, and
  Khudanpur}]{snyder2018x}
Snyder, D.; Garcia-Romero, D.; Sell, G.; Povey, D.; and Khudanpur, S. 2018.
\newblock X-vectors: Robust dnn embeddings for speaker recognition.
\newblock In \emph{2018 IEEE international conference on acoustics, speech and
  signal processing (ICASSP)}, 5329--5333. IEEE.

\bibitem[{Song et~al.(2020)Song, Wu, Qian, He, and Loy}]{song2020everybody}
Song, L.; Wu, W.; Qian, C.; He, R.; and Loy, C.~C. 2020.
\newblock Everybody's talkin': Let me talk as you want.
\newblock \emph{arXiv preprint arXiv:2001.05201}.

\bibitem[{Song et~al.(2018)Song, Zhu, Li, Wang, and Qi}]{song2018talking}
Song, Y.; Zhu, J.; Li, D.; Wang, X.; and Qi, H. 2018.
\newblock Talking face generation by conditional recurrent adversarial network.
\newblock \emph{arXiv preprint arXiv:1804.04786}.

\bibitem[{Suwajanakorn, Seitz, and
  Kemelmacher-Shlizerman(2017)}]{suwajanakorn2017synthesizing}
Suwajanakorn, S.; Seitz, S.~M.; and Kemelmacher-Shlizerman, I. 2017.
\newblock Synthesizing obama: learning lip sync from audio.
\newblock \emph{ACM Transactions on Graphics (ToG)}, 36(4): 1--13.

\bibitem[{Thies et~al.(2020)Thies, Elgharib, Tewari, Theobalt, and
  Nie{\ss}ner}]{thies2020neural}
Thies, J.; Elgharib, M.; Tewari, A.; Theobalt, C.; and Nie{\ss}ner, M. 2020.
\newblock Neural voice puppetry: Audio-driven facial reenactment.
\newblock In \emph{ECCV}, 716--731. Springer.

\bibitem[{Van~der Maaten and Hinton(2008)}]{van2008visualizing}
Van~der Maaten, L.; and Hinton, G. 2008.
\newblock Visualizing data using t-SNE.
\newblock \emph{Journal of machine learning research}, 9(11).

\bibitem[{Vaswani et~al.(2017)Vaswani, Shazeer, Parmar, Uszkoreit, Jones,
  Gomez, Kaiser, and Polosukhin}]{vaswani2017attention}
Vaswani, A.; Shazeer, N.; Parmar, N.; Uszkoreit, J.; Jones, L.; Gomez, A.~N.;
  Kaiser, {\L}.; and Polosukhin, I. 2017.
\newblock Attention is all you need.
\newblock In \emph{Advances in neural information processing systems},
  5998--6008.

\bibitem[{Vougioukas, Petridis, and Pantic(2019)}]{vougioukas2019realistic}
Vougioukas, K.; Petridis, S.; and Pantic, M. 2019.
\newblock Realistic speech-driven facial animation with gans.
\newblock \emph{International Journal of Computer Vision}, 1--16.

\bibitem[{Wang et~al.(2020)Wang, Wu, Song, Yang, Wu, Qian, He, Qiao, and
  Loy}]{wang2020mead}
Wang, K.; Wu, Q.; Song, L.; Yang, Z.; Wu, W.; Qian, C.; He, R.; Qiao, Y.; and
  Loy, C.~C. 2020.
\newblock Mead: A large-scale audio-visual dataset for emotional talking-face
  generation.
\newblock In \emph{ECCV}, 700--717. Springer.

\bibitem[{Wang et~al.(2021)Wang, Li, Ding, Fan, and Yu}]{wang2021audio2head}
Wang, S.; Li, L.; Ding, Y.; Fan, C.; and Yu, X. 2021.
\newblock Audio2Head: Audio-driven One-shot Talking-head Generation with
  Natural Head Motion.
\newblock \emph{IJCAI}.

\bibitem[{Wang et~al.(2022)Wang, Li, Ding, and Yu}]{wang2022one}
Wang, S.; Li, L.; Ding, Y.; and Yu, X. 2022.
\newblock One-shot talking face generation from single-speaker audio-visual
  correlation learning.
\newblock In \emph{Proceedings of the AAAI Conference on Artificial
  Intelligence}, volume~36, 2531--2539.

\bibitem[{Wang and Tu(2020)}]{wang2020rethinking}
Wang, W.; and Tu, Z. 2020.
\newblock Rethinking the value of transformer components.
\newblock In \emph{Proceedings of the 28th International Conference on
  Computational Linguistics}, 6019--6029.

\bibitem[{Wang et~al.(2004)Wang, Bovik, Sheikh, and Simoncelli}]{wang2004image}
Wang, Z.; Bovik, A.~C.; Sheikh, H.~R.; and Simoncelli, E.~P. 2004.
\newblock Image quality assessment: from error visibility to structural
  similarity.
\newblock \emph{IEEE transactions on image processing}, 13(4): 600--612.

\bibitem[{Wiles, Koepke, and Zisserman(2018)}]{wiles2018x2face}
Wiles, O.; Koepke, A.; and Zisserman, A. 2018.
\newblock X2face: A network for controlling face generation using images,
  audio, and pose codes.
\newblock In \emph{ECCV}, 670--686.

\bibitem[{Wu et~al.(2021)Wu, Jia, Wang, Dou, Duan, and Deng}]{wu2021imitating}
Wu, H.; Jia, J.; Wang, H.; Dou, Y.; Duan, C.; and Deng, Q. 2021.
\newblock Imitating arbitrary talking style for realistic audio-driven talking
  face synthesis.
\newblock In \emph{Proceedings of the 29th ACM International Conference on
  Multimedia}, 1478--1486.

\bibitem[{Yang et~al.(2019)Yang, Bender, Le, and Ngiam}]{yang2019condconv}
Yang, B.; Bender, G.; Le, Q.~V.; and Ngiam, J. 2019.
\newblock Condconv: Conditionally parameterized convolutions for efficient
  inference.
\newblock \emph{Advances in Neural Information Processing Systems}, 32.

\bibitem[{Yi et~al.(2020)Yi, Ye, Zhang, Bao, and Liu}]{yi2020audio}
Yi, R.; Ye, Z.; Zhang, J.; Bao, H.; and Liu, Y.-J. 2020.
\newblock Audio-driven talking face video generation with learning-based
  personalized head pose.
\newblock \emph{arXiv preprint arXiv:2002.10137}.

\bibitem[{Yin et~al.(2022)Yin, Zhang, Cun, Cao, Fan, Wang, Bai, Wu, Wang, and
  Yang}]{yin2022styleheat}
Yin, F.; Zhang, Y.; Cun, X.; Cao, M.; Fan, Y.; Wang, X.; Bai, Q.; Wu, B.; Wang,
  J.; and Yang, Y. 2022.
\newblock Styleheat: One-shot high-resolution editable talking face generation
  via pretrained stylegan.
\newblock \emph{arXiv preprint arXiv:2203.04036}.

\bibitem[{Yu et~al.(2018)Yu, Fernando, Ghanem, Porikli, and
  Hartley}]{yu2018face}
Yu, X.; Fernando, B.; Ghanem, B.; Porikli, F.; and Hartley, R. 2018.
\newblock Face super-resolution guided by facial component heatmaps.
\newblock In \emph{ECCV}, 217--233.

\bibitem[{Yu and Porikli(2017{\natexlab{a}})}]{yu2017face}
Yu, X.; and Porikli, F. 2017{\natexlab{a}}.
\newblock Face hallucination with tiny unaligned images by transformative
  discriminative neural networks.
\newblock In \emph{AAAI}.

\bibitem[{Yu and Porikli(2017{\natexlab{b}})}]{yu2017hallucinating}
Yu, X.; and Porikli, F. 2017{\natexlab{b}}.
\newblock Hallucinating very low-resolution unaligned and noisy face images by
  transformative discriminative autoencoders.
\newblock In \emph{CVPR}, 3760--3768.

\bibitem[{Zhang et~al.(2021{\natexlab{a}})Zhang, Ni, Fan, Li, Zeng, Budagavi,
  and Guo}]{zhang20213d}
Zhang, C.; Ni, S.; Fan, Z.; Li, H.; Zeng, M.; Budagavi, M.; and Guo, X.
  2021{\natexlab{a}}.
\newblock 3d talking face with personalized pose dynamics.
\newblock \emph{IEEE Transactions on Visualization and Computer Graphics}.

\bibitem[{Zhang et~al.(2021{\natexlab{b}})Zhang, Zhao, Huang, Zeng, Ni,
  Budagavi, and Guo}]{zhang2021facial}
Zhang, C.; Zhao, Y.; Huang, Y.; Zeng, M.; Ni, S.; Budagavi, M.; and Guo, X.
  2021{\natexlab{b}}.
\newblock FACIAL: Synthesizing Dynamic Talking Face with Implicit Attribute
  Learning.
\newblock In \emph{ICCV}, 3867--3876.

\bibitem[{Zhang et~al.(2021{\natexlab{c}})Zhang, Li, Ding, and
  Fan}]{zhang2021flow}
Zhang, Z.; Li, L.; Ding, Y.; and Fan, C. 2021{\natexlab{c}}.
\newblock Flow-Guided One-Shot Talking Face Generation With a High-Resolution
  Audio-Visual Dataset.
\newblock In \emph{CVPR}, 3661--3670.

\bibitem[{Zhou et~al.(2019)Zhou, Liu, Liu, Luo, and Wang}]{zhou2019talking}
Zhou, H.; Liu, Y.; Liu, Z.; Luo, P.; and Wang, X. 2019.
\newblock Talking face generation by adversarially disentangled audio-visual
  representation.
\newblock In \emph{AAAI}, volume~33, 9299--9306.

\bibitem[{Zhou et~al.(2021)Zhou, Sun, Wu, Loy, Wang, and Liu}]{zhou2021pose}
Zhou, H.; Sun, Y.; Wu, W.; Loy, C.~C.; Wang, X.; and Liu, Z. 2021.
\newblock Pose-controllable talking face generation by implicitly modularized
  audio-visual representation.
\newblock In \emph{CVPR}, 4176--4186.

\bibitem[{Zhou et~al.(2020)Zhou, Han, Shechtman, Echevarria, Kalogerakis, and
  Li}]{zhou2020makelttalk}
Zhou, Y.; Han, X.; Shechtman, E.; Echevarria, J.; Kalogerakis, E.; and Li, D.
  2020.
\newblock MakeltTalk: speaker-aware talking-head animation.
\newblock \emph{ACM Transactions on Graphics (TOG)}, 39(6): 1--15.

\bibitem[{Zhu et~al.(2021)Zhu, Luo, Wang, Zheng, and He}]{zhu2021deep}
Zhu, H.; Luo, M.-D.; Wang, R.; Zheng, A.-H.; and He, R. 2021.
\newblock Deep audio-visual learning: A survey.
\newblock \emph{International Journal of Automation and Computing}, 1--26.

\end{thebibliography}




\end{document}